\documentclass[sigconf, screen, authorversion]{acmart}
\AtBeginDocument{%
  }

\setcopyright{acmlicensed}
\copyrightyear{2025}
\acmYear{2025}
\acmDOI{XXXXXXX.XXXXXXX}
\acmConference[ACM Multimedia '25]{}{October 27--31,
  2025}{Dublin, Ireland}

\usepackage{booktabs}
\usepackage{array}
\usepackage{graphicx}
\usepackage{tabularx}
\usepackage{newpxtext}
\usepackage{multirow}
\usepackage{adjustbox}
\usepackage{cleveref}
\usepackage{colortbl}
\makeatletter
\newcommand{\ie}{\emph{i.e.}\@ifnextchar.{\!\@gobble}{}}
\newcommand{\eg}{\emph{e.g.}\@ifnextchar.{\!\@gobble}{}}
\newcommand{\etc}{etc\@ifnextchar.{}{.\@}}
\makeatother

\begin{document}

\title{SVC 2025: the First Multimodal Deception Detection Challenge}

\author{Xun Lin}
\authornote{Equal Contribution.}
\affiliation{%
  \institution{Great Bay University}
  \city{Dongguan, Guangdong}
  \country{China}
}
\email{lxlinxun@gmail.com}

\author{Xiaobao Guo}
\authornotemark[1]
\affiliation{%
  \institution{Nanyang Technological University}
  \country{Singapore}}
\email{xiaobao001@e.ntu.edu.sg}

\author{Taorui Wang}
\authornotemark[1]
\affiliation{%
  \institution{Great Bay University}
  \city{Dongguan, Guangdong}
  \country{China}
}
\email{25B351018@stu.hit.edu.cn}

\author{Yingjie Ma}
\authornotemark[1]
\affiliation{%
  \institution{Great Bay University}
  \city{Dongguan, Guangdong}
  \country{China}
}
\email{murinj2248@outlook.com}

\author{Jiajian Huang}
\authornotemark[1]
\affiliation{%
  \institution{Great Bay University}
  \city{Dongguan, Guangdong}
  \country{China}}
\email{jiajian_huang_cs@163.com}

\author{Jiayu Zhang}
\authornotemark[1]
\affiliation{%
  \institution{Great Bay University}
  \city{Dongguan, Guangdong}
  \country{China}}
\email{qmmcxm@protonmail.com}

\author{Junzhe Cao}
\authornotemark[1]
\affiliation{%
  \institution{Great Bay University}
  \city{Dongguan, Guangdong}
  \country{China}
}
\email{caojunzhe@buaa.edu.cn}

\author{Zitong Yu}
\authornote{Corrsponding Author}
\affiliation{%
  \institution{Great Bay University}
  \city{Dongguan, Guangdong}
  \country{China}
}
\email{zitong.yu@ieee.org}

\renewcommand{\shortauthors}{Xun et al.}

\begin{abstract}
Deception detection is a critical task in real-world applications such as security screening, fraud prevention, and credibility assessment. While deep learning methods have shown promise in surpassing human-level performance, their effectiveness often depends on the availability of high-quality and diverse deception samples. Existing research predominantly focuses on single-domain scenarios, overlooking the significant performance degradation caused by domain shifts. To address this gap, we present the SVC 2025 Multimodal Deception Detection Challenge, a new benchmark designed to evaluate cross-domain generalization in audio-visual deception detection. Participants are required to develop models that not only perform well within individual domains but also generalize across multiple heterogeneous datasets. By leveraging multimodal data, including audio, video, and text, this challenge encourages the design of models capable of capturing subtle and implicit deceptive cues. Through this benchmark, we aim to foster the development of more adaptable, explainable, and practically deployable deception detection systems, advancing the broader field of multimodal learning. By the conclusion of the workshop competition, a total of 21 teams had submitted their final results. Our baseline is released at~\hyperlink{https://github.com/Redaimao/MMDD2025}{MMDD2025}.
\end{abstract}

\begin{teaserfigure}
\centering
  \includegraphics[width=\textwidth]{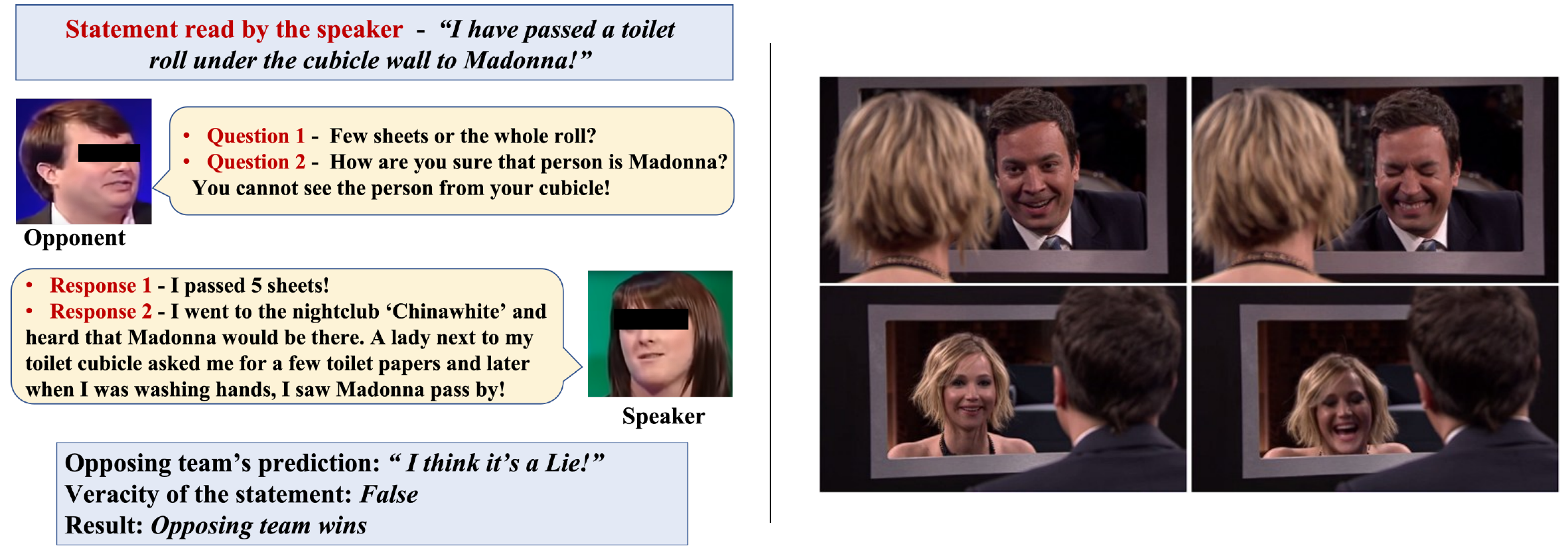}
    \caption{Examples of deceptive actions. The left side shows a game show scenario from~\citeauthor{guo2023audio}, illustrating a typical deceptive round. The right image displays selected frames showcasing truthful and deceptive behaviors from video clips provided by~\citeauthor{soldner2019box}.}
  \Description{Example of deceptive action.}
  \label{fig:teaser}
\end{teaserfigure}


\maketitle

\section{Introduction}

Deception detection plays a crucial role in accurately assessing truthfulness and identifying deceptive behaviors, having pivotal applications in many
fields such as credibility assessment in business, multimedia anti-fraud, and custom security~\cite{abouelenien2016detecting,glancy2011computational,joudaki2014using}.
With its significant intention, deception detection remains inherently difficult. As human bias towards assuming truthfulness, human precision remains around 54\%~\cite{bond2006accuracy}, slightly above chance. To discover a better performance and a less time-consuming method, researchers have increasingly explored automated approaches that combine advances in computer vision, natural language processing, and deep learning for deception detection. Recently, deep learning methods have demonstrated their credibility, achieving comparable or surpassing human detection even in some complex tasks~\cite{silver2016mastering,silver2018general,brown2020language}.

The performance of AI models in deception detection is heavily reliant on the availability of authentic and effective deception samples from the real world. While present models provide satisfactory results, fewer studies have explored the cross-domain issue, despite the presence of significant domain shifts in public deception detection datasets~\cite{soldner2019box,gupta2019bag,csen2020multimodal,lloyd2019miami}. The generalizability of the models is critical for practical applications. Therefore, such domain shifts need to be investigated in order to develop deception detection models that can be generalized across different contexts.

To obtain a precise detection, multimodal deception detection (MMDD) came into being. Following~\cite{guo2024benchmarking,guo2023audio}, MMDD is a typical subtle visual computing task, aiming to detect imperceptible and deceptive clues from audio-visual scenarios. But general performance of cross-domain deception detection is unsatisfactory because it is challenging to reduce the domain gap between each dataset. In response to this, we propose the first Multimodal deception detection (MMDD) challenge - \textbf{SVC 2025}~\footnote{https://sites.google.com/view/svc-mm25}, aiming to bring together researchers and developers to advance the field of multimodal learning by detecting deception through the integration of multiple modalities such as audio, video, and text. The competition encourages innovation in building robust AI models that can accurately identify deceptive behaviors by leveraging various features from these modalities.
Participants are required to submit their developed model, checkpoints and well-explained source code, accompanied by a paper describing their proposed methodologies and the achieved results. Only contributions that meet the predetermined requirements, terms and conditions are eligible for participation. The organisers do not engage in active participation themselves, but instead undertake a re-evaluation of the findings of the systems submitted to challenge. The ranking of the submitted models depends on three metrics: accuracy, error rate, and F1-score for ranking on the test dataset split, with accuracy being the primary metric. 

In summary, the main contributions and novelties of this challenge are listed as follows:
\begin{itemize}
    \item We first introduce a new benchmark for the cross-domain audio-visual deception detection challenge. We present a comprehensive benchmark that evaluates the generalization capacity of AI models using audio and visual modalities across multiple domains in the deception detection task. Consequently, the SVC 2025 challenge requires models not only to be able to detect in a single domain, but also across multiple domains following three distinct domain sampling strategies,~\ie, domain simultaneous, domain alternating, and domain-by-domain.
    \item We introduce a novel protocol that enhances the model's generalization capability by evaluating performance across multiple domains, rather than limiting to a single-domain setting as in prior work. This better reflects real-world deployment conditions, where models must adapt to unseen or shifting data distributions, and therefore encourages the development of more robust and widely applicable solutions.
\end{itemize}
\section{Related Works}
\subsection{Deception Detection Approaches}
The research on using behavioral cues for deception has gradually become active over the past few decades. Among the studied behavioral cues, verbal and nonverbal cues were preferred as humans may behave differently between lying and telling the truth.
Traditional deception detection is often a contact-based method. It assesses whether someone is telling the truth or not by monitoring physiological responses like skin conductance and heart rate~\cite{synnott2015review,li2024deception,javaid2022eeg}. Study~\cite{depaulo2003cues} has revealed that, in general, people who tell lies are less forthcoming and less convincing than those who tell the truth. Liars usually talk about fewer details and make fewer spontaneous corrections. They also sound less involved but more vocally tense. Through the study, the researchers statistically found that liars often press their lips, repeat words, raise their chins, and show less genuine smiles. The results show that some behavioral cues do potentially appear in deception and are even more pronounced when liars are more motivated to cheat.

\subsection{Multimodal Deception Detection}
Recent advances in deception detection have increasingly incorporated both verbal and non-verbal cues, leveraging multimodal data to enhance detection performance.
\citeauthor{gogate2017deep} proposed a deep model that incorporated the audio cues with visual and text modalities to improve the accuracy of deception prediction. \citeauthor{karimi2018toward} explored deceptive cues from RGB images and raw audio in an end-to-end manner. \citeauthor{wu2018deception} utilized several types of features, including micro-expression and IDT (Improved Dense Trajectory) features from RGB images, MFCC (Mel-frequency Cepstral Coefficients) features from the audio, and transcripts.

Despite significant progress, there still exists the challenge of cross-domain generalization in multimodal deception detection. Previous works have mainly focused on optimizing unimodal or fusion methods within a single domain, without considering the domain shift when the system is applied to different environments or populations. For example, models trained on controlled lab data may not generalize well to real-world settings, where the recording conditions and communication styles may introduce significant variability.

In this challenge, we aim to follow the benchmark~\cite{guo2024benchmarking} for cross-domain generalization performance on the widely used audio-visual deception detection datasets, which is crucial for evaluating and improving the robustness of deception detection models in different scenarios. Establishing such a challenge will provide clearer comparisons, highlight model weaknesses, and guide the development of systems across different domains.
\section{Challenge Corpora}


The competition employed three datasets as training datasets: Real-life Deception Detection (Real-life Trial)~\cite{csen2020multimodal,perez2015deception}, Bag-of-Lies~\cite{gupta2019bag}, and the Miami University Deception Detection Database (MU3D)~\cite{lloyd2019miami}, as shown in~\cref{table: dataset comparison table}. All participants must sign an agreement before accessing the datasets on their original platforms. Competition organizers will not provide raw data directly to participants. Instead, extracted OpenFace features, affect features from pretrained models, and Mel spectrograms (generated using PyTorch) are provided. These features do not contain any identifiable information.

\begin{table*}[]
\centering
\label{table:dataset comparison table}
\begin{adjustbox}{max width=0.8\textwidth}
    \begin{tabular}{ccccccc}
    \toprule
    \multirow{2}{*}{\textbf{Dataset}} & \multirow{2}{*}{\textbf{\#Subjects}} & \multicolumn{4}{c}{\textbf{Samples}} & \multirow{2}{*}{\textbf{Scenario}} \\ 
     &    & \multicolumn{1}{c}{\textbf{Total}} & \multicolumn{1}{c}{\textbf{Deceptive}} & \multicolumn{1}{c}{\textbf{Truthful}} & \textbf{Deception/Truth Ratio} &  \\
     \midrule
    Real Life Trials\cite{perez2015deception}  & 56 & \multicolumn{1}{c}{121} & \multicolumn{1}{c}{61} & \multicolumn{1}{c}{60} & 1.02 & Court-room \\
    
    Bag of Lies\cite{gupta2019bag}   & 35 & \multicolumn{1}{c}{325} & \multicolumn{1}{c}{162} & \multicolumn{1}{c}{163} & 0.99 & Lab \\
    MU3D\cite{lloyd2019miami}   & 80 & \multicolumn{1}{c}{320} & \multicolumn{1}{c}{160} & \multicolumn{1}{c}{160} & 1 & Lab \\

    \midrule
    Box of Lies\cite{soldner2019box}  & 26 & \multicolumn{1}{c}{1049} & \multicolumn{1}{c}{862} & \multicolumn{1}{c}{187} & 4.61 & Gameshows \\

    \bottomrule
    \end{tabular}
\end{adjustbox}
\caption{Multimodal deception detection datasets in challenge}
\vspace{-1.5em}
\label{table: dataset comparison table}
\end{table*}

The Real-life Trial Deception Dataset contains 121 video clips from real courtroom proceedings, averaging 28 seconds each. These clips include famous cases like the Jodi Arias trial, exoneration testimonies from "The Innocence Project," and defendant statements from crime-related TV episodes, featuring statements by defendants or witnesses. With 61 clips labeled deceptive and 60 truthful based on trial outcomes (guilty verdicts, not-guilty verdicts, and exonerations), the dataset covers 21 female and 35 male speakers aged 16–60. It also provides manually verified crowdsourced transcripts (8,055 words, including fillers/repetitions) and annotations of 9 non-verbal gesture categories (e.g., facial expressions, hand movements) using the MUMIN coding scheme.

The Bag-of-Lies dataset contains 325 annotated recordings from 35 unique subjects, including 162 deceptive and 163 truthful samples. It integrates four modalities: video capturing facial and body expressions via smartphone, audio of speech descriptions, gaze tracking data with fixation points and pupil metrics collected using Gazepoint GP3, and EEG signals from a 13-channel Emotiv EPOC+ headset sampled at 128 Hz. During collection, participants freely described 6-10 distinct images, choosing spontaneously to lie or tell the truth per image. Recording durations ranged from 3.5 to 42 seconds.

The Miami University Deception Detection Database (MU3D) consists of 320 videos featuring 80 distinct participants of different races. Each participant generated four videos: a positive truth describing someone they genuinely liked, a negative truth describing someone they genuinely disliked, a positive lie falsely portraying a disliked person as liked, and a negative lie falsely portraying a liked person as disliked. The videos were collected in a laboratory setting where participants responded to standardized prompts for 45 seconds.

We use the Box of Lies game show dataset~\cite{soldner2019box} for Stage 1 evaluation. This dataset contains 1,049 annotated utterances extracted from 25 publicly available video clips of The Tonight Show Starring Jimmy Fallon. The total video footage spans 2 hours and 24 minutes. It documents interactions between host Jimmy Fallon and 26 guests, including 6 males and 20 females, capturing both truthful and deceptive behaviors. Data collection involved extracting YouTube videos, segmenting conversational turns via ELAN software, and annotating multimodal behaviors such as facial expressions, head movements, and gaze according to the MUMIN coding scheme. Verbal content was transcribed via Amazon Mechanical Turk and manually verified for accuracy. Each utterance carries a veracity label, including 862 deceptive and 187 truthful instances. Linguistic features were extracted from transcripts, and non-verbal behaviors (e.g., smile frequency) were quantified as temporal percentages.




\section{Evaluation Metrics}
This competition employs three primary evaluation metrics: accuracy, error rate, and F1-score. Among these, \textbf{accuracy serves as the principal ranking criterion} for participant submissions. All metrics are computed based on binary classification outcomes where:
\begin{itemize}
    \item Truthful samples are labeled as \textbf{1}
    \item Deceptive samples are labeled as \textbf{0}
\end{itemize}

\begin{enumerate}
    \item \textbf{Accuracy}: Proportion of correctly classified samples
    \[
    \text{Accuracy} = \frac{TP + TN}{TP + TN + FP + FN}
    \]
    where:
    \begin{itemize}
        \item $TP$: True positives (correctly predicted deceptive)
        \item $TN$: True negatives (correctly predicted truthful)
        \item $FP$: False positives (truthful misclassified as deceptive)
        \item $FN$: False negatives (deceptive misclassified as truthful)
    \end{itemize}
    
    \item \textbf{Error Rate}: Complement of accuracy representing misclassification frequency
    \[
    \text{Error Rate} = 1 - \text{Accuracy} = \frac{FP + FN}{TP + TN + FP + FN}
    \]
    
    \item \textbf{F1-score}: Harmonic mean of precision and recall
    \[
    \text{Precision} = \frac{TP}{TP + FP}, \quad
    \text{Recall} = \frac{TP}{TP + FN}
    \]
    \[
    F1 = 2 \times \frac{\text{Precision} \times \text{Recall}}{\text{Precision} + \text{Recall}}
    \]
\end{enumerate}

Predictions are generated as continuous scores $\in [0,1]$ representing the probability of a sample being deceptive. For metric computation, these scores are thresholded at 0.5:
\begin{align*}
\text{Predicted class} = 
\begin{cases} 
0 \text{ (deceptive)} & \text{if score} \geq 0.5 \\
1 \text{ (truthful)} & \text{if score} < 0.5 
\end{cases}
\end{align*}
The evaluation process requires participants to submit prediction files containing sample paths and corresponding scores, formatted as:
\begin{verbatim}
SJ_BOL_EP3_lie_4 0.14431
\end{verbatim}




\section{Baseline model}

We employ a cross-domain audio-visual deception detection method as the baseline model~\cite{guo2024benchmarking}, extracting face frame features via ResNet18, acquiring behavioral features (AUs, gaze, and affect) using OpenFace and EmotionNet, and extracting Mel spectrograms for audio via OpenSmile or processing raw waveforms with Wave2Vec. Features from each modality are encoded and fused through modules like linear layers or Transformer before being fed into classifiers. 

Cross-domain generalization strategies include single-to-single and multi-to-single scenarios. The latter introduces three sampling strategies: 
\begin{itemize}
\item Domain-Simultaneous mixes samples from multiple source domains per batch to learn domain-invariant features; 
\item Domain-Alternating samples from one source domain batch by batch to capture domain specifics; 
\item Domain-by-Domain trains on source domains sequentially but may overfit to single-domain features. 
\end{itemize}

\begin{figure}[htbp]
    \centering
    \includegraphics[width=0.5\textwidth]{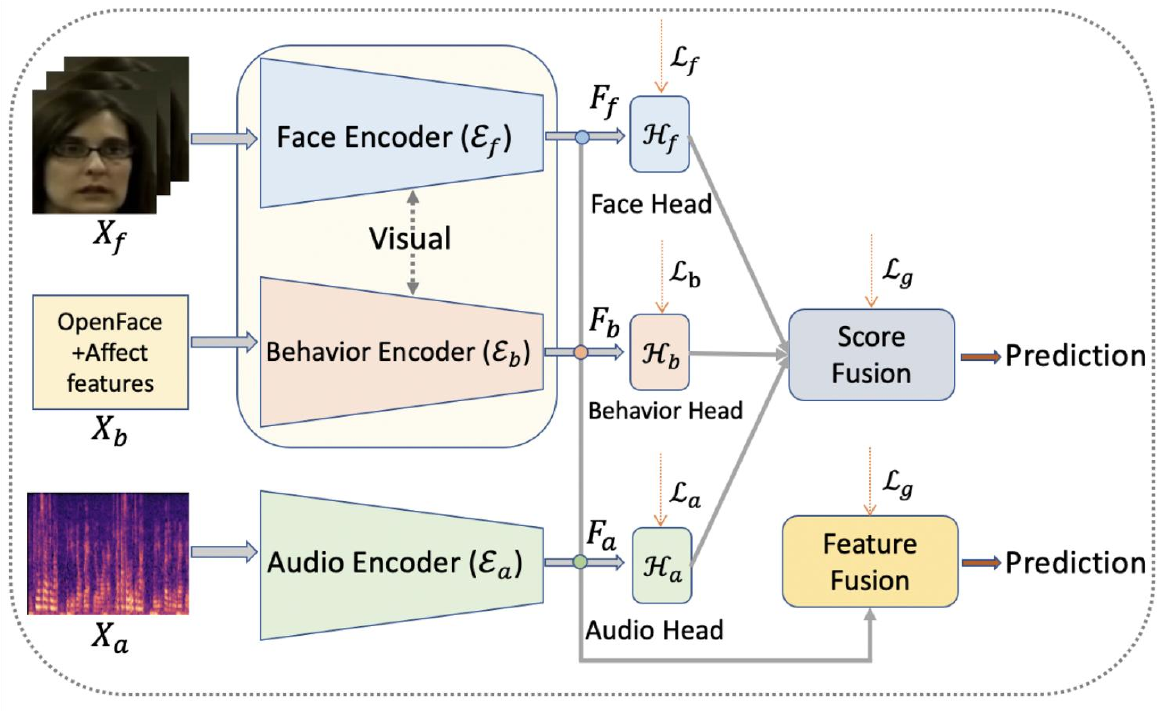}
    \caption{baseline model}
    \label{fig:baseline}
\end{figure}

The model architecture is depicted in Fig ~\ref{fig:baseline}.

The MM-IDGM algorithm is proposed to align cross-domain gradient directions by maximizing gradient inner products between modality encoders. It dynamically adjusts learning rates based on the "Fish" algorithm and optimizes with previous unimodal losses to enhance multi-to-single generalization. 

The Attention-Mixer fusion method combines MLP-Mixer and self-attention layers. It uses unimodal MLP layers to learn single-modality feature dynamics, self-attention layers to explore intra-feature associations, and cross-modal MLP layers to capture inter-modal interactions. With layer normalization and multi-head self-attention, six attention mixer layers are stacked: input features are projected, processed through multiple layers, and mean-pooled to output, enhancing intra- and inter-modal interactions for improved fusion performance.





\section{Participation}

A total of 21 teams submitted their results by the end of the workshop competition~\footnote{https://codalab.lisn.upsaclay.fr/competitions/22162\#results}. The results of phase 2 are shown in Table.~\ref{tab:leaderboard}. We will introduce the approaches of some teams in the following part.

\begin{table}[htbp]
\centering

\setlength{\tabcolsep}{4pt}
\renewcommand{\arraystretch}{1.2}
\begin{tabularx}{\linewidth}{@{}c>{\centering\arraybackslash}X>{\centering\arraybackslash}p{1.6cm}>{\centering\arraybackslash}p{1.6cm}>{\centering\arraybackslash}p{1.6cm}@{}}
\toprule
\textbf{\#} & \textbf{Team} & \textbf{ACC} & \textbf{F1} & \textbf{ERR} \\
\midrule
1 & \textbf{Glenn\_xxy} & \cellcolor{yellow!40}62.437396 & 43.890274  & \cellcolor{yellow!40}37.562604 \\
2 & \textbf{BigHandsome} & \cellcolor{gray!20}60.434057  & \cellcolor{yellow!40}56.987296  & \cellcolor{gray!20}39.565943  \\
3 & \textbf{aim\_whu} & 58.931553   & \cellcolor{gray!20}45.333333  & 41.068447   \\
\bottomrule
\end{tabularx}
\caption{Challenge Results.\textcolor{yellow!40}{\rule{6pt}{6pt}} indicates the \textbf{best} result; \textcolor{gray!20}{\rule{6pt}{6pt}} indicates the \textbf{second-best} result. Note that only the top-3 teams are shown here.}
\label{tab:leaderboard}

\end{table}


\subsection{Team Glenn\_xxy}
\begin{figure}[htbp]
    \centering
    \includegraphics[width=0.5\textwidth]{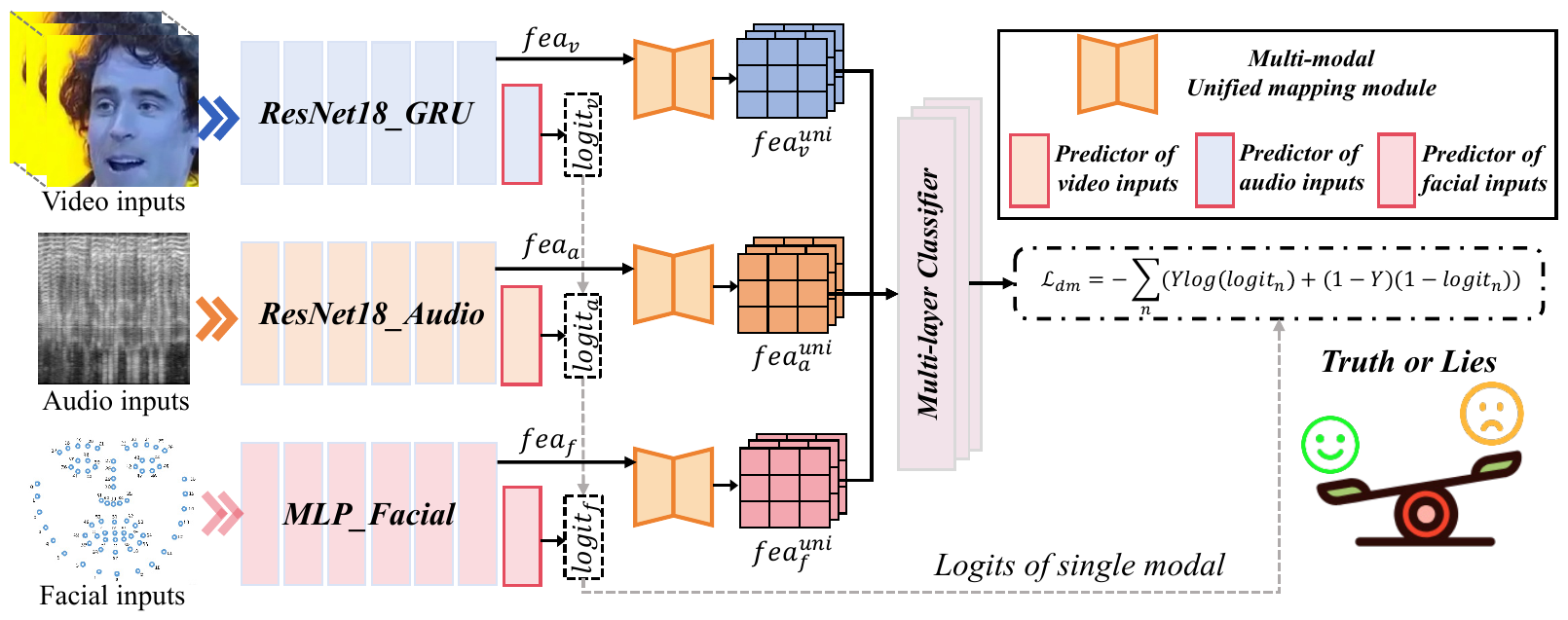}
    \caption{Network framework of Team Glenn\_xxy}
    \label{fig:first}
\end{figure}

Team Glenn\_xxy proposed LCUNet, which employs three modality-specific branches (ResNet for audio and fusion, MLP for video) to extract initial features and logits from input images. Instead of simple 1x1 convolution for alignment, it introduces modality-specific projection networks to map these features into a unified space. These projection networks downsample the feature dimensions to half the original size and then upsample back to the original scale, maintaining spatial resolution. The unified features are concatenated and processed by a deeper classifier to produce a fused logit. The final prediction combines the individual modality logits and the fused logit, leveraging both single-modality and fused discriminative information. The framework of proposed LCUNet is shown in Fig \ref{fig:first}.

\subsection{Team BigHandsome}
\begin{figure}[htbp]
    \centering
    \includegraphics[width=0.5\textwidth]{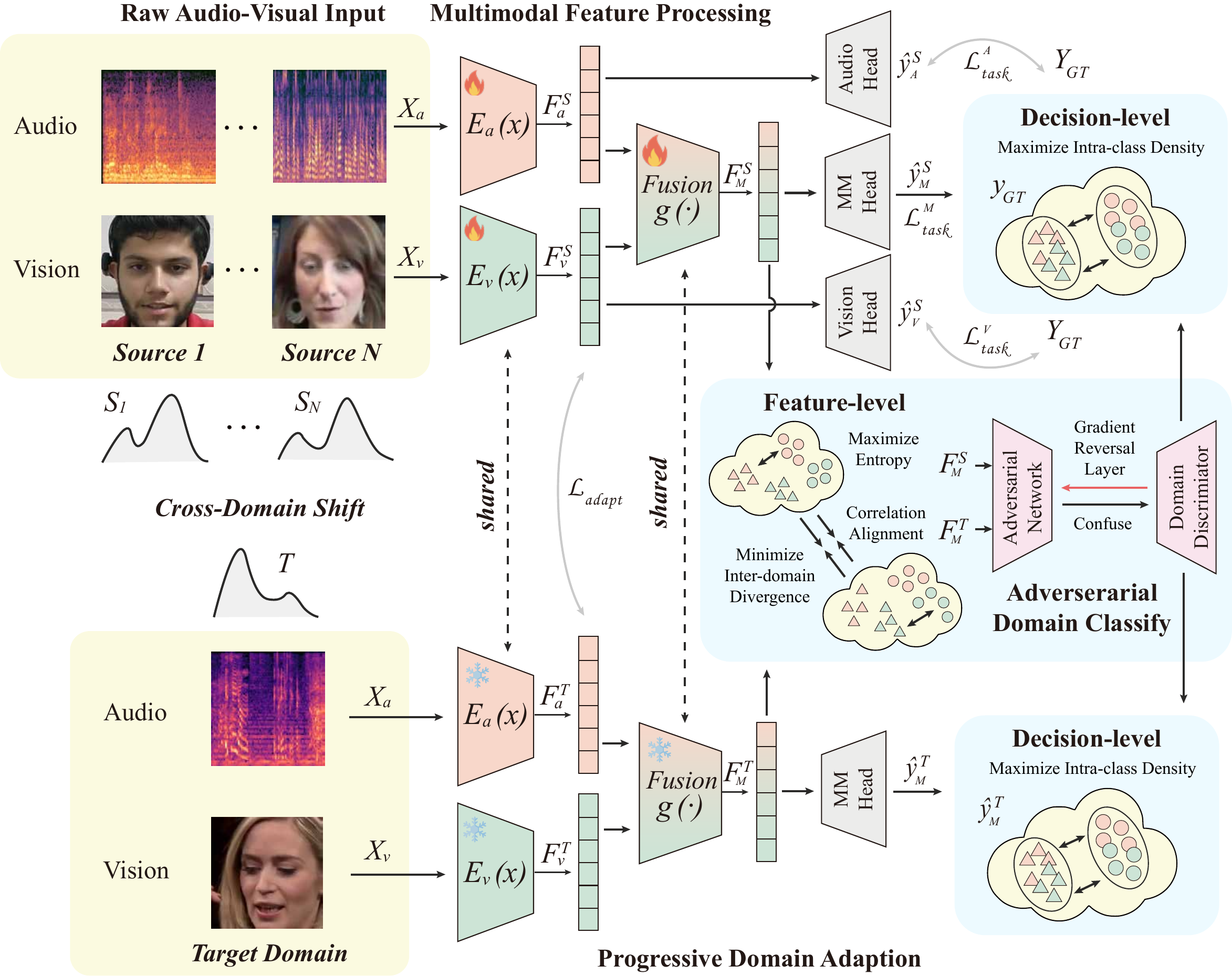}
    \caption{Network framework of Team BigHandsome}
    \label{fig:second}
\end{figure}

Team BigHandsome, noting the significant domain gap across different datasets, employed a multi-loss framework to align training and testing features. Specifically, datasets are processed domain-by-domain. Multiple domain generalization loss functions are utilized simultaneously: CORAL loss for correlation alignment, Maximum Mean Discrepancy (MMD) loss for density divergence, Entropy Maximization (ATM) loss to confuse the domain discriminator, and Adversarial loss (based on CDAN+E principles) combined with a Gradient Reversal Layer for adversarial domain adaptation during backpropagation. The model schematic is presented in Fig \ref{fig:second}.

\subsection{Team aim\_whu}
\begin{figure}[htbp]
    \centering
    \includegraphics[width=0.5\textwidth]{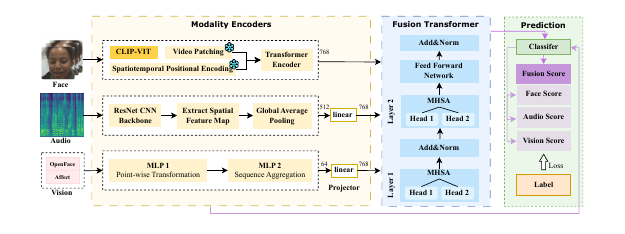}
    \caption{Network framework of Team aim\_whu}
    \label{fig:third}
\end{figure}

Recognizing that ViT's global attention mechanism can more accurately capture fleeting and widespread cues in deceptive behaviors, Team aim\_whu selected Vision Transformer (ViT) as the facial feature extractor, replacing the baseline ResNet. Key training improvements include: Merging heterogeneous deception detection datasets to enhance sample diversity and generalizability; Adopting a Cosine Annealing LR scheduler (replacing StepLR) to achieve smooth learning rate decay and stable convergence; Implementing a differential learning rate strategy—applying a smaller learning rate to the pre-trained ViT backbone to preserve its knowledge, while using higher rates for other components. The model schematic is illustrated in Fig \ref{fig:third}.

\section{Discussion and Future Prospect }
This challenge marks the first multimodal deception detection competition, serving as a valuable benchmark for the research community, fostering the development and application of multimodal data integration and fusion methods. Participants investigated visual, acoustic, and behavioral features etc, highlighting the potential of combining complementary multimodal information for superior deception detection. One notable strength of this challenge was its emphasis on intricate, real-world scenarios, leveraging diverse modalities and multiple datasets. This encourages innovative fusion strategies and the development of practical deception detection methods. However, one limitation was the short timeframe, which constrained participants from fully leveraging the raw multimodal data for deeper methodological exploration, particularly in areas such as temporal dynamics and end-to-end learning. Future challenge iterations could use longer timelines and bigger datasets for better models. Moreover, integrating large-scale pretrained models (e.g., multimodal foundation models) and introducing explainability tools would not only enhance predictive performance but also improve interpretability and trustworthiness of deception detection methods. These directions open exciting opportunities for advancing the field both in research and application.
\section{Conclusion}
As the competition demonstrates, we can see that multimodal deception detection still has great potential for development. We hope that both the new challenge and baseline code, as well as the methodologies and outcomes of the participating teams, will serve as a helpful stepping stone for researchers who are interested in multimodal deception detection, leading to real-world applications.

\begin{acks}

\end{acks}

\bibliographystyle{ACM-Reference-Format}
\bibliography{ref}










\end{document}